\documentclass[12pt]{article}
\usepackage[left=1in,top=1in,right=1in,bottom=1in,headheight=16pt]{geometry}

\usepackage{cite}
\usepackage[square,numbers]{natbib}
\usepackage{amsmath,amssymb,amsfonts}
\usepackage{algorithmic}
\usepackage{graphicx}
\usepackage{textcomp}
\usepackage{xcolor}
\usepackage{authblk}

\begin{document}

\title{Simulations evaluating resampling methods for causal discovery: ensemble performance and calibration\footnote{This work was supported by funding from NCRR 1UL1TR002494-01, 1R01MH116156-01A1, and 1R03MH117254-01A1 to EK.}
}

\author[1]{Erich Kummerfeld}
\author[2]{Alexander Rix}
\affil[1]{Institute for Health Informatics, University of Minnesota}
\affil[2]{Department of biostatistics, University of Michigan}

\maketitle

\begin{abstract}
Causal discovery can be a powerful tool for investigating causality when a system can be observed but is inaccessible to experiments in practice. Despite this, it is rarely used in any scientific or medical fields. One of the major hurdles preventing the field of causal discovery from having a larger impact is that it is difficult to determine when the output of a causal discovery method can be trusted in a real-world setting. Trust is especially critical when human health is on the line.

In this paper, we report the results of a series of simulation studies investigating the performance of different resampling methods as indicators of confidence in discovered graph features. We found that subsampling and sampling with replacement both performed surprisingly well, suggesting that they can serve as grounds for confidence in graph features. We also found that the calibration of subsampling and sampling with replacement had different convergence properties, suggesting that one's choice of which to use should depend on the sample size.
\end{abstract}

\section{Introduction}
Many scientific disciplines seek to use controlled experiments to build, contradict, or confirm causal models in their respective fields. It is often not possible to conduct controlled experiments, however, for a variety of practical and ethical reasons. Fields that study human health are especially vulnerable to these limitations. In such circumstances, researchers must rely on observational data. Causal discovery methods provide a way to learn causal information from observational data, and so one might expect causal discovery methods to be heavily utilized in the medical sciences, but at the time of writing this is not the case. One major reason for this is that, unlike in some other machine learning domains, it is difficult to determine whether or not the output of a causal discovery method is trustworthy in a real-world setting. Prediction algorithms can be evaluated relatively simply by, for example, measuring the area under the receiver operator curve (AUC) on a holdout sample. However causal graphs do not lend themselves to such an evaluation. An alternative approach is needed.

In this paper we investigate one approach to evaluating the performance of causal discovery methods: resampling. Resampling methods such as the bootstrap, and less commonly the jackknife, have been used heavily in other areas of statistics and machine learning, and it seems natural to extend this approach to causal discovery. Many of the known statistical properties of the bootstrap and jackknife do not apply in an obvious way to the causal discovery setting, however, leaving open the problem of evaluating resampling as a method for evaluating causal discovery applications.

\subsection{Related work}

We are only aware of one previous publication investigating resampling calibration in the context of causal discovery algorithms. Naeini, Jabbari, and Cooper \citep{fattaneh} examined the the calibration of directed edges in a bootstrapped version of the Really Fast Causal Inference \citep{colombo2012learning} algorithm and found that they were generally well calibrated. In this paper, we look at both bootstrapping and jackknifing, use a different simulation setup, use a different base search algorithm, and evaluate all edge types.

Most of the previous work in bootstrapping graphical models has been done in the broader field of learning bayesian networks. Friedman, Goldszmidt, and Wyner successfully used bootstrapped hill-climbing, showed that bootstrapped hill-climbing's edge probabilities were well calibrated, and found that bootstrapped hill-climbing substantially outperformed the unbootstrapped version \citep{friedman1999data,friedman1999application}. Steck and Jaakola showed that when using hill-climbing on bootstrapped data, the Bayesian Information Criterion (BIC) score is biased towards overly dense graphs, sometimes severely, and derived a correction \citep{steck2004bias}. They also showed that the jackknife is not affected by this bias. It is unclear at this time whether these results extend to causal discovery.

\subsection{Methods}
This paper reports the results of a series of simulations that we ran as a first step towards a better understanding of resampling in the context of causal discovery. Three different types of simulations were run: one simulation randomly generated causal models with random structures and random linear Gaussian parameters, and two simulations used existing expert models meant to be representative of medical phenomena. In these simulations, data were generated from a causal graphical model, and then repeatedly resampled, resulting in large collections of data sets. Each of these resampled data sets was then analyzed with a causal discovery algorithm, producing a corresponding large collection of graphs. We then calculated the proportion of times each pair of variables had a particular kind of relationship, as represented by the edge (or lack thereof) connecting them, within this collection of graphs. Finally, we evaluated these proportions as forecasts of whether the edge was present or not in the data generating model, and calculated both the full Brier score of the forecasts as well as their reliability (calibration).

\subsection{Results}
In our simulations, subsampling (jackknifing) and resampling with replacement (bootstrapping) performed well both in terms of overall Brier score and in terms of reliability, even at smaller sample sizes, for linear Gaussian networks. Both procedures had poor Brier scores with data simulated from expert models over categorical variables. The jackknife also had poor calibration on the expert models, while the bootstrap had excellent calibration on one expert model and poor calibration on the other. The two procedures appeared to have different convergence properties when the sample size was varied. The bootstrap had better performance than the jackknife in most, but not all, circumstances.

\subsection{Outline}
This paper proceeds as follows. In section \ref{background}, a brief background is provided on the material covered in this paper: causal discovery, resampling, and the Brier score, a standard method for evaluating calibration. Section \ref{simulation} provides the specifications for the simulations we ran. Section \ref{results} provides the outcomes of those simulations, including evaluations of both the ensemble accuracy and calibration of the resampling techniques. The paper concludes with section \ref{discussion}, which discusses the results, their implications, and their limitations, and considers future directions for this line of research.

\section{Background}
\label{background}
\subsection{Causal discovery}

\begin{figure}[htbp]
\centerline{\includegraphics[width=0.4\columnwidth]{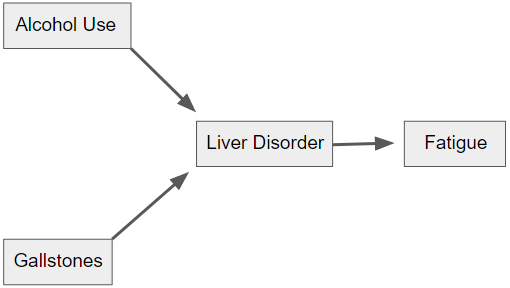}}
\caption{Example of a causal graph. In this example, Alcohol Use and Gallstones cause Liver Disorder, which in turn causes Fatigue.}
\label{example}
\end{figure}

Causal discovery algorithms estimate the structure of a causal model \citep{pearl2000causality}, represented as a graph, using data generated from that model \citep{spirtes2000causation}. Figure \ref{example} shows an example of a causal graph. Because of the general structure of the problem, these algorithms have a common high-level form: they take data as input, and produce a graph as output. Beyond that, however, they vary considerably. The field is too large to review here, so we will only discuss the algorithm used in this investigation; for recent introductory papers on causal discovery, see \citep{eberhardt2017introduction} or \citep{malinsky2018causal}.

Greedy Equivalence Search (GES) \citep{chickering2002learning,Meek:1995:CIC:2074158.2074204,ramsey2017million} is a fast and scalable causal discovery algorithm that is correct when there are no causal cycles or unmeasured common causes. It accomplishes this by performing an intelligent search within the space of all causal models to find the model that optimizes a model fit statistic chosen by the user. The most common fit statistic used for this purpose is the Bayesian Information Criterion (BIC) score:

$$\text{BIC} = -2 \log L + k \log n$$

Where $L$ is the likelihood of the data given the model, $n$ is the sample size, and $k$ is the number of free parameters in the model. $L$ is computed after fitting the model's free parameters to the data with a maximum likelihood estimate (MLE) procedure, so it is the highest likelihood that the model can assign to the data. The implementation of GES used in this paper includes a \emph{penalty discount} parameter as well, which acts as a tuning parameter for preferring more complex or more simple models by modifying the strength of the second term in the BIC score.  This modified BIC score's second term is thus $d k \log n$, where $d$ is the user-assigned value for the penalty discount parameter.

\subsection{Resampling}
The bootstrap was introduced by Efron \citep{efron1979} and comes in two primary forms: the parametric bootstrap fits a (parametric) model to the data and samples from the model, and the non parametric bootstrap instead samples from the empirical distribution function by sampling with replacement from the data. We use only non parametric bootstrapping in this paper. In statistics, the bootstrap can be used for many things, including estimating the standard error of an estimator, and constructing confidence intervals. The delete $d$ jackknife is a similar procedure, where one takes the original sample and randomly deletes $d$ observations from the data. $d$ is usually chosen on the order of $\sqrt{n}$, as small numbers can lead to inconsistency when the statistic of interest is not smooth. We chose to include the jackknife due to \citep{steck2004bias}, where the jackknife performed well and generally chose sparser models than the bootstrap in the the context of Bayesian Networks.

In machine learning the bootstrap has primarily been used as a way to improve the performance of prediction algorithms via bootstrap aggregating, or bagging. Bagging was introduced by Breiman \citep{breiman1996bagging} as way to improve the predictive performance of learning algorithms. Bagging involves generating $m$ bootstrap replicates of the original data and employing a rule to condense the $m$ models into a single value. In the case of bagging classifiers, \citep{breiman1996bagging} suggests generating a frequency vector and classifying based off the most frequent observation, which is the procedure we used in this paper.
In the context of learning Bayesian networks, \citep{friedman1999application} and \citep{friedman1999data} define the probability of a graph feature (e.g. individual edges and adjacencies) as

$$ p_n(f) = \frac{1}{|D_n|}\sum_{D_n} I(f \in \hat{G}(D_n)) $$
where $D_n$ is any possible data set of size $n$ sampled from the Bayesian network $B$, and $\hat{G}(D_n)$ is the graph learned from $D_n$.  They use the bootstrap to estimate $p$
$$ p_n(f) \approx \frac{1}{m} \sum_m I(f \in \hat{G}(D_n^m)) $$
where $D_n^m$ is the $m$-th bootstrapped data set. As \citep{hastie01esl} notes, the resulting feature frequencies from the bagging procedure are not probabilities. We treat these quantities only as frequencies with evaluable calibration.

Resampling adds a few levels of complexity to typical causal discovery. First, there are different resampling procedures to choose from. Second, running the learning algorithm on resampled data can be biased \citep{steck2004bias}, which may require adjustment in other parts of the procedure in order to correct for this bias. Third, there are multiple ways to use the resampling graphs, such as aggregating them into a single object or looking at the distribution of their features. We explore both of these uses in this paper.

\subsection{Brier score and reliability}
In order to evaluate the calibration of these resampling techniques for GES, we treat the resampling frequencies as \emph{forecasts}, and evaluate their forecasting ability using the Brier score\citep{brier1950verification}, a standard tool for evaluating the performance of probabilistic forecasts for binary events. The Brier score is simply the mean squared error between the predicted probability and the observed outcomes:

$$\text{Brier Score} = \frac{1}{n}\sum_{s=1}^{n}(f_s-o_s)^2$$
where $n$ is the sample size, $f_s$ is the forecast probability of the event in sample $s$, and $o_s$ is the observed outcome in sample $s$, with 0 meaning the event did not occur and 1 meaning the event did occur. Since this is a measure of error, larger values are worse: the best Brier score possible is 0, while the worst score possible is 1.

We also used one of the components that the Brier score can be decomposed into\citep{murphy1973new}, typically called \emph{reliability} or \emph{calibration}. Here we will call it reliability, in order to distinguish it from the more general concept of calibration. This component evaluates how close each forecast is to the actual frequency with which the event occurs when that forecast was made:

$$\text{Reliability} = \frac{1}{n}\sum_{k=1}^{K}n_k(f_k-\bar{o}_k)^2$$

where $K$ is the number of distinct forecasts, $n_k$ is the number of samples where forecast $k$ was made, $f_k$ is the forecast probability of the event when forecast $k$ is made, and $\bar{o}_k$ is the observed frequency of the event occurring among samples where forecast $k$ was made. Like with the overall Brier score, smaller reliability scores are preferred. The optimal reliability is 0, while the worst possible reliability is 1. Ferro and Fricker\citep{ferro2012bias} showed that the standard decomposition of the Brier score is biased, and provided a modified version of reliability with less bias. We used both versions, and found the difference to be minor. We report only the bias-corrected reliability in this paper.

\section{Simulation Specifications}
\label{simulation}
\subsection{Simulation design}
We implemented a series of simulations to evaluate the performance of resampling as a method for determining confidence in individual applications of causal discovery. The overall structure of these simulations is as follows.

First, a graphical model is selected and its parameters are assigned values. For the linear Gaussian models, the graphical structures were randomly generated, and the parameter values are randomly drawn from defined distributions. For the expert model simulations, both the graphical structures and parameter values were determined by the expert model. Second, data are randomly generated from the model, up to a specified sample size. For all simulations, sample sizes were varied from 100 to 1000 in increments of 100. Both steps are then repeated, resulting in a collection of matched $\langle\text{graph},\text{data} \rangle$ pairs. Finally, each data set is analyzed, and the output of that analysis is compared against the matched data-generating graph.

\subsection{Algorithm and resampling specifications}

While we varied our overall analysis techniques, we only made use of one pre-existing causal discovery method, the Greedy Equivalence Search (GES) \citep{chickering2002learning,Meek:1995:CIC:2074158.2074204,ramsey2017million}. GES was always run with the Bayesian Information Criterion (BIC) and a penalty discount of 2, and all other parameters were left at their default values. We used R \citep{R} to automate our application of GES to all simulated data sets. For all simulations we used the implementation of GES found in Rix's open source R package ``Causality''\footnote{https://github.com/tzimiskes/causality}, as we found it to be faster than alternative methods for running GES in R.

We used two different resampling methods. First, based on the bootstrap method, we used sampling with replacement to the full original sample size. Second, based on the jackknife method, we used sampling without replacement to 90\% of the original sample size. In both cases, 200 resampled data sets would be made from the original data set. For the rest of the paper we will refer to these procedures simply as ``bootstrap'' and ``jackknife'' procedures.

The resampling methods were used in two different ways. For a direct comparison with standard GES, we used the resampling methods to produce an ensemble estimate of the data generating graph via a bagging procedure. The ensemble estimate is calculated by taking the set of output graphs from the set of resampled data sets, and letting the graphs vote on the relationship between every pair of variables, with the highest vote winning. This produces a partially directed graph as output, but is not guaranteed to be a pattern, or even to be acyclic.

We also used the resampling methods to produce forecasts of the edges of the data generating graph. The forecasts were calculated similar to that of the ensemble estimates, except rather than producing a single graph as output, the output is a table of the proportion of votes each relationship got for each variable pair. These proportions can then be treated as forecasts of the probability that that variable pair has that particular relationship.

\subsection{Model specifications}

\paragraph{Linear Gaussian models}
We produced 500 independent $\langle\text{graph},\text{data} \rangle$ pairs of linear Gaussian models and data sets at each sample size. The data was generated using the graphical user interface of the open source Tetrad software package\footnote{http://www.phil.cmu.edu/tetrad/}, version 6.6.0. All graphs were generated with 100 variables and 100 edges. Their structures were randomly selected with the \textit{random forward} process. Each graph was assigned parameters independently in the following way. Each variable is a weighted sum of the value of its parents in the graph and its own independent noise term. Each variable's independent noise term has a Normal distribution with mean 0 and a variance drawn uniformly between 1 and 3. The weight of the independent noise term is always 1, and each incoming edge has a weight drawn from SplitUniform(-1.5, -0.5; 0.5, 1.5).

\paragraph{Expert models}

We used the Child \citep{spiegelhalter1992learning} and Hepar2 \citep{onisko2003probabilistic} expert models. Data from both was generated using the bnlearn R package \citep{bnlearn,R}. 100 data sets were generated from each model at each sample size.

\section{Results}
\label{results}

\subsection{Linear Gaussian simulation results}
\paragraph{Graph estimation performance}

\begin{figure}[htbp]
\centerline{\includegraphics[width=.5\columnwidth]{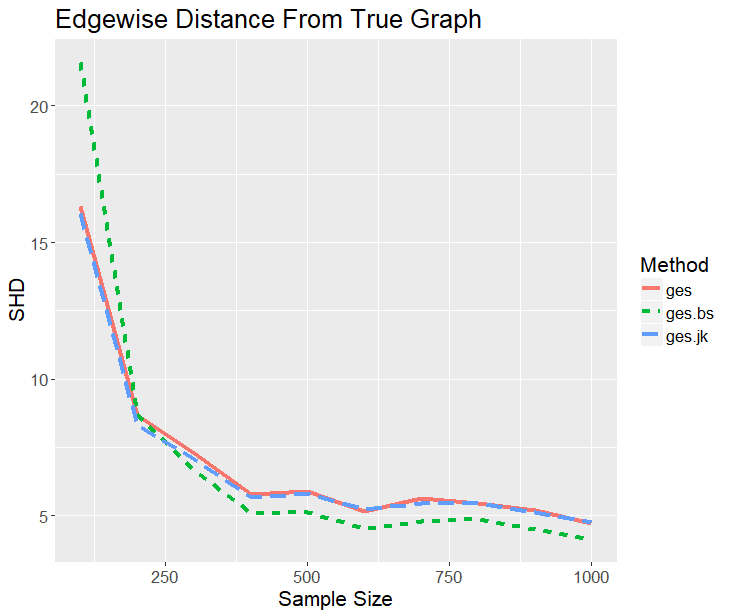}}
\caption{Average Structural Hamming Distance of algorithm output on raw data compared to data generating model. Although overall performance is good relative to the size of the graphs, errors are still present.}
\label{SHD}
\end{figure}

To evaluate the impact of using resampling ensembles on GES, we calculated the Structural Hamming Distance (SHD) \citep{tsamardinos2006max} of the standard GES algorithm as well as the jackknife and bootstrap ensemble methods. SHD is a numeric summary of the edgewise distance of the graph estimate from the data generating graph: each missing edge, extra edge, or wrongly oriented edge adds 1 point to the SHD. Figure \ref{SHD} shows a plot of the SHD by sample size in our simulations. Since in these simulations the data generating graphs all had 100 nodes and 100 edges, these errors are fairly small, suggesting that our linear Gaussian data generation procedure was relatively easy for GES to learn, even at lower sample sizes.

Perhaps because the learning problem was relatively easy, the difference between the three methods were small. The jackknife ensemble approach had similar SHD to standard GES, while the bootstrap ensemble approach performed worse at the lowest sample size of 100, but by sample size 300 was slightly outperforming the other two methods. We also calculated and plotted the adjacency recall and precision, and arrowhead recall and precision, for these methods, but omit them here as the results were captured by the SHD plot. Those plots are available on request.

\begin{figure}[htbp]
\centerline{\includegraphics[width=.5\columnwidth]{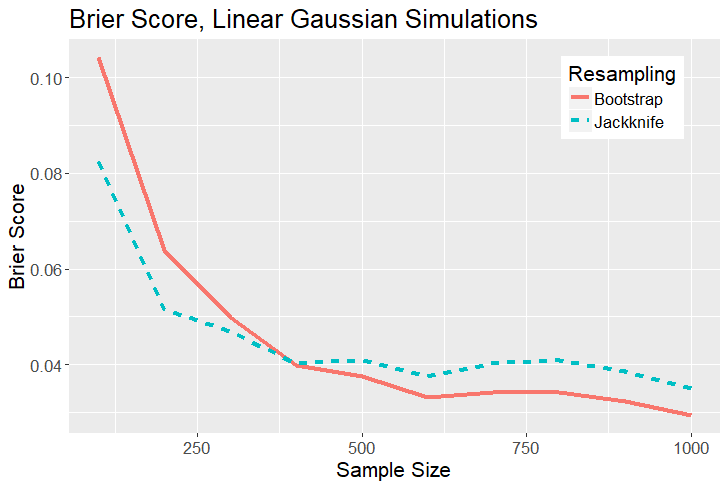}}
\caption{Plot of the Brier scores of forecasts from the bootstrap and jackknife procedures.}
\label{brier}
\end{figure}

\paragraph{Forecasting performance}
To evaluate the bootstrap and jackknife procedures as methods for forecasting the probability that a given edge was present in the data generating model, we calculated the Brier score and reliability of each method. Figure \ref{brier} shows a plot of the Brier scores of the bootstrap and jackknife procedures across all sample sizes. This plot mimics the ensemble performance plot, as the bootstrap procedure had worse Brier scores at lower sample sizes, but better scores at higher sample sizes. Compared to the ensemble plot, the crossover point also occurs at a higher sample size, as the bootstrap procedure does not catch up to the jackknife procedure until the sample size is at least 400. Overall, though, both methods performed reasonably well, with their Brier scores at or below .05 for most sample sizes.

\begin{figure}[htbp]
\centerline{\includegraphics[width=.5\columnwidth]{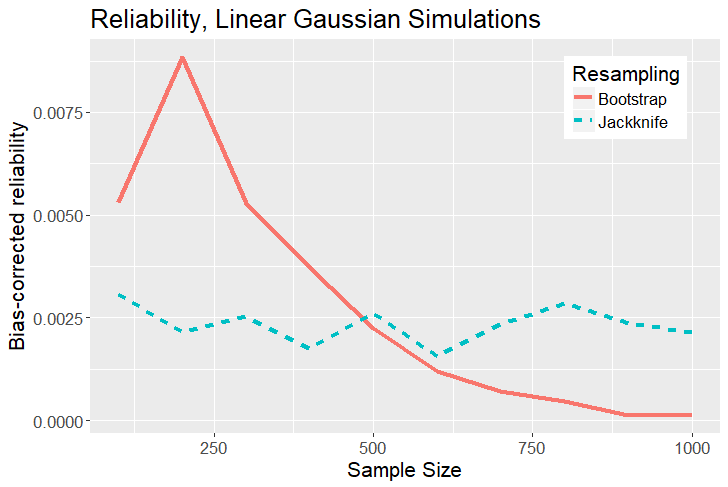}}
\caption{Plot of the bias-corrected reliability (calibration) of the bootstrap and jackknife procedures.}
\label{reliability}
\end{figure}

Figure \ref{reliability} shows a plot of the bias-corrected reliability of the bootstrap and jackknife procedures across all sample sizes. Overall both methods performed well, as the bootstrap procedure never rose above 0.01, corresponding to forecasts that are typically within 10\% of the observed event frequency, and the jackknife procedure never rose above 0.003, corresponding to forecasts that are typically within 5\% of the observed event frequency. This plot again mimics the ensemble performance plot and Brier score plot, however the difference between the methods appears more pronounced, with the bootstrap procedure performing proportionally much worse than the jackknife at sample sizes below 400, but much better at sample sizes above 800. Also of note, the bootstrap's reliability appears to converge quickly towards 0 as the sample sizes increases, while the jackknife's reliability appears independent of sample size, staying near 0.002 at all sample sizes.

\begin{figure}[htbp]
\centerline{\includegraphics[width=.5\columnwidth]{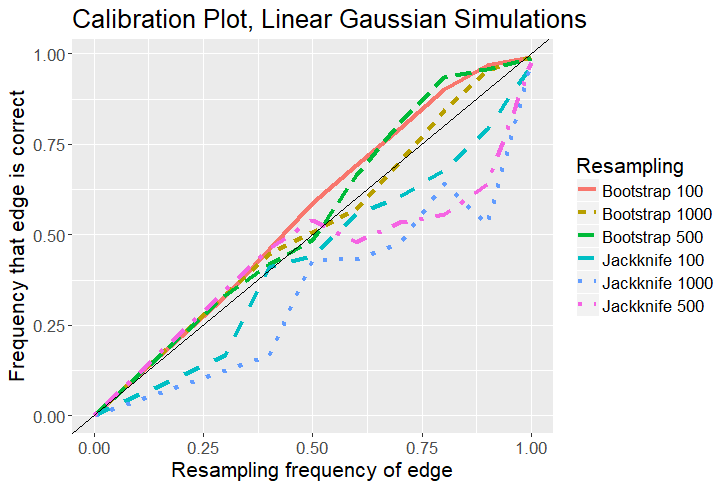}}
\caption{Comparison of forecast accuracy to actual accuracy for bootstrap and jackknife procedures on simulated linear Gaussian data sets with sample sizes 100, 500, and 1000. The diagonal black line represents perfect forecasts.}
\label{ss100-500-1000-cal}
\end{figure}

Figure \ref{ss100-500-1000-cal} shows a plot of the actual frequency of correct edges against the forecast frequency of those edges being correct, for both methods at sample sizes 100, 500, and 1000. Visual inspection of this plot also suggests that both methods are well calibrated, but reveals some additional differences in their performance. Forecasts from the bootstrap procedure appear to underestimate the actual frequency, while forecasts from the jackknife procedure appear to overestimate the actual frequency.

\subsection{Expert model simulation results}

\paragraph{Hepar2}
The Hepar2 model was developed by Onisko \citep{onisko2003probabilistic} as a model of the causes and effects of liver disorders. It was originally developed to aid in the diagnosis of liver disorders. It has 70 nodes and 123 edges. The model was downloaded from the bnlearn Bayesian Network Repository, and the data were simulated from the model using the bnlearn package \citep{bnlearn}.

\begin{figure}[htbp]
\centerline{\includegraphics[width=.5\columnwidth]{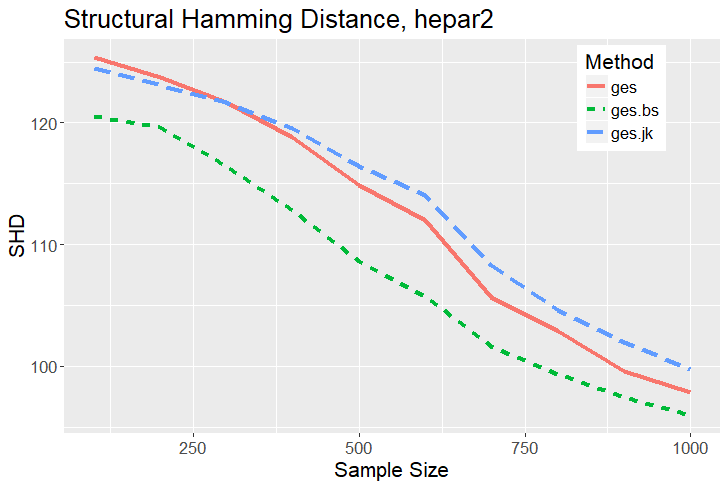}}
\caption{Comparison of Structural Hamming Distance for bootstrap and jackknife procedures on data sets simulated from the Hepar2 expert model, for sample sizes 100 to 1000. Lower values are better.}
\label{hepar2shd}
\end{figure}

Figure \ref{hepar2shd} shows the accuracy performance of GES, GES with a bootstrap ensemble, and GES with a jackknife ensemble on the Hepar2 simulated data. All methods performed poorly on all sample sizes tested: SHD ranged between 90 and 130, which is a large number of errors given the size of the hepar2 model. All methods appear to still be in the process of converging by sample size 1000. The bootstrap ensemble outperformed both raw GES and the jackknife ensemble. An inspection of the more detailed recall and precision statistics of these methods revealed that most errors were recall errors. All methods had fairly high precision, meaning that identified edges were typically correct, however all methods also had very low recall, meaning that the methods typically did not find most of the edges in the hepar2 model.

\begin{figure}[htbp]
\centerline{\includegraphics[width=.5\columnwidth]{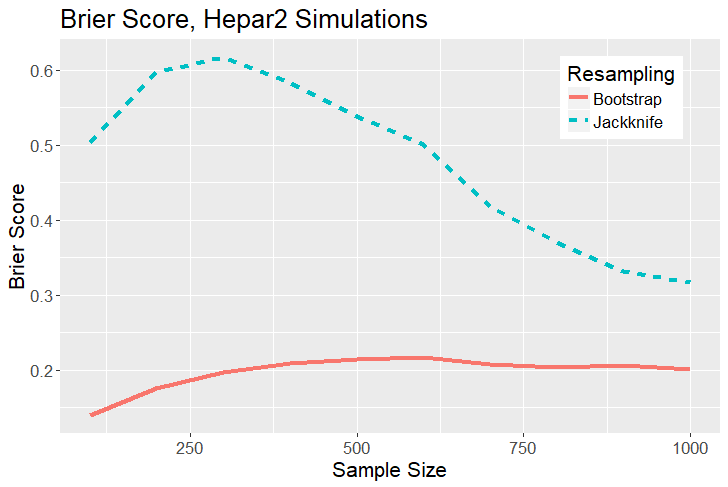}}
\caption{Comparison of Brier score for bootstrap and jackknife procedures on data sets simulated from the Hepar2 expert model, for sample sizes 100 to 1000. Lower values are better.}
\label{hepar2brier}
\end{figure}

\begin{figure}[htbp]
\centerline{\includegraphics[width=.5\columnwidth]{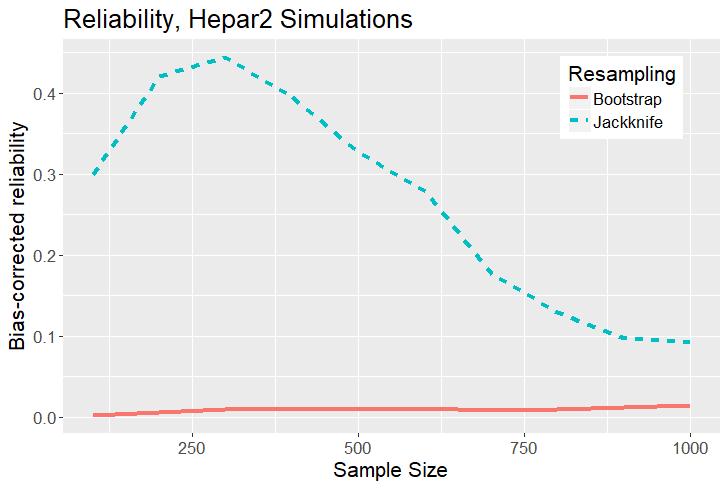}}
\caption{Comparison of reliability for bootstrap and jackknife procedures on data sets simulated from the Hepar2 expert model, for sample sizes 100 to 1000. Lower values are better.}
\label{hepar2rel}
\end{figure}

Figures \ref{hepar2brier} and \ref{hepar2rel} show plots of the resampling procedures' Brier scores and reliability, respectably. The bootstrap procedure performed notably better in both cases.  While neither method did as well as we might want on total Brier score at any sample size, the bootstrap procedure's forecasts were still highly reliable at all sample sizes. Somewhat interestingly, neither procedure appears to be converging monotonically towards the ideal Brier score or reliability in the sample sizes tested. For both Brier score and reliability, the jackknife procedure is actually performing worse as the sample size grows, until 300 samples, where it turns and improves continuously up through sample size 1000. For the bootstrap procedure, performance slowly gets worse until around sample size 500, where it seems to stabilize.

\begin{figure}[htbp]
\centerline{\includegraphics[width=.5\columnwidth]{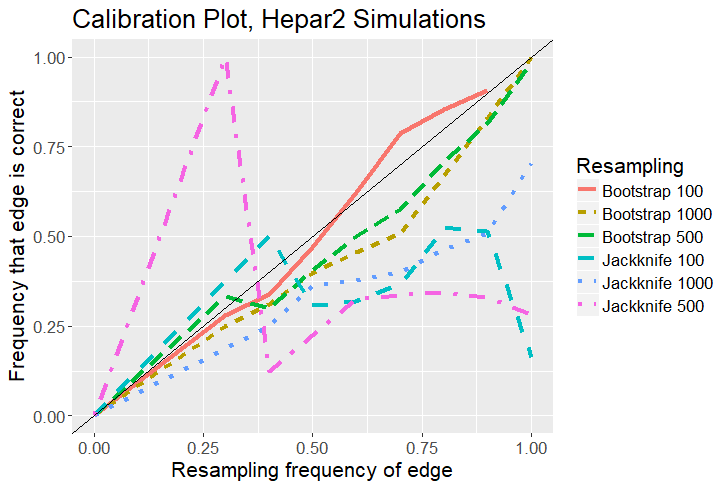}}
\caption{Comparison of forecast accuracy to actual accuracy for bootstrap and jackknife procedures on data sets simulated from the Hepar2 expert model with sample size 500. The diagonal black line represents perfect forecasts.}
\label{hepar2_cal}
\end{figure}

Figure \ref{hepar2_cal} shows a calibration plot for both procedures at sample sizes 100, 500, and 1000. Both methods were typically overestimating the likelihood of edges being correct, with the jackknife procedure making more extreme errors of this kind than the bootstrap procedure. The jackknife's calibration performance is also much noisier than the bootstrap's calibration performance.

\paragraph{child}

The Child model \citep{spiegelhalter1992learning} models the effects of birth asphyxia and the outcomes of some related medical exams. It has 20 nodes and 25 edges. The model was downloaded from the bnlearn Bayesian Network Repository, and the data were simulated from the model using the bnlearn package \citep{bnlearn}.

\begin{figure}[htbp]
\centerline{\includegraphics[width=.5\columnwidth]{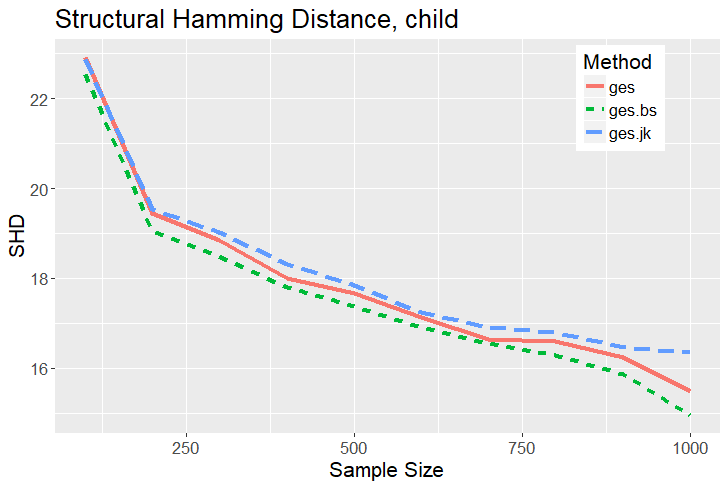}}
\caption{Comparison of Structural Hamming Distance for bootstrap and jackknife procedures on data sets simulated from the child expert model, for sample sizes 100 to 1000. Lower values are better.}
\label{childshd}
\end{figure}

Figure \ref{childshd} shows the accuracy performance of GES, GES with a bootstrap ensemble, and GES with a jackknife ensemble on the Child simulated data. As with the Hepar2 simulation, all methods performed poorly for the sample sizes tested, and appear to still be converging at sample size 1000. The bootstrap ensemble again appears to have the best performance, although the difference is small.

\begin{figure}[htbp]
\centerline{\includegraphics[width=.5\columnwidth]{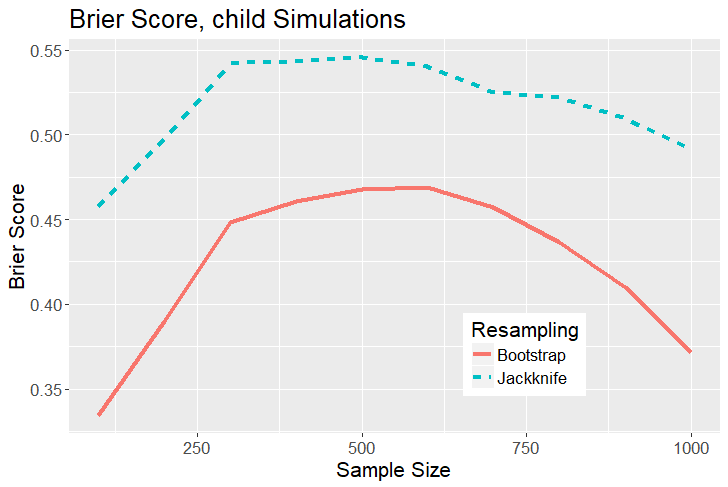}}
\caption{Comparison of Brier score for bootstrap and jackknife procedures on data sets simulated from the Hepar2 expert model, for sample sizes 100 to 1000. Lower values are better.}
\label{childbrier}
\end{figure}

\begin{figure}[htbp]
\centerline{\includegraphics[width=.5\columnwidth]{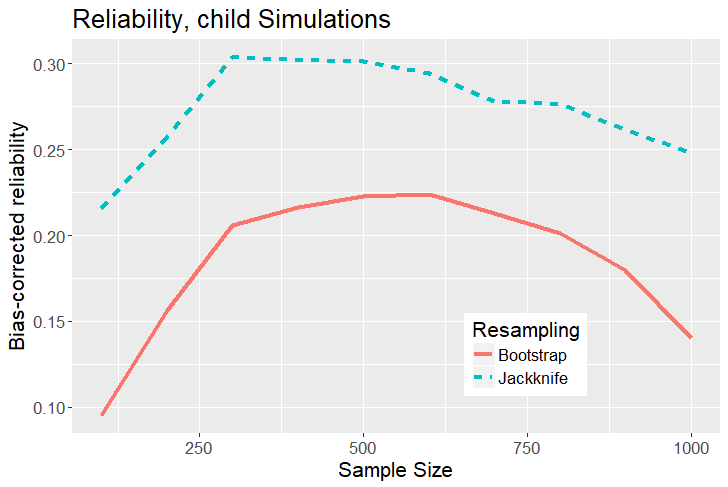}}
\caption{Comparison of reliability for bootstrap and jackknife procedures on data sets simulated from the Hepar2 expert model, for sample sizes 100 to 1000. Lower values are better.}
\label{childrel}
\end{figure}

Figures \ref{childbrier} and \ref{childrel} show plots of the Brier scores and reliability, respectably, for both resampling procedures. As with the Hepar2 simulation, Brier scores were poor overall, the jackknife procedure had poor reliability, and the bootstrap procedure outperformed the jackknife procedure on both measures at all sample sizes. Unlike the Hepar2 simulation, the bootstrap did not have strong reliability in the Child simulation. Also unlike the Hepar2 simulation, the Brier scores and reliability for the bootstrap and jackknife procedures changed in similar ways as the sample size increased from 100 to 1000. Both traced a concave shape, with performance degrading from sample size 100 to approximately sample size 500, where it began to improve in an accelerated manner up to sample size 1000, the maximum tested.

\begin{figure}[htbp]
\centerline{\includegraphics[width=.5\columnwidth]{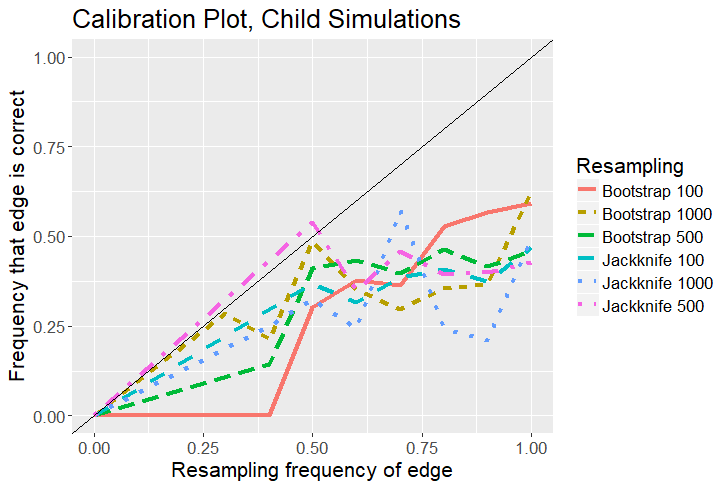}}
\caption{Comparison of forecast accuracy to actual accuracy for bootstrap and jackknife procedures on data sets simulated from the Child expert model with sample size 500. The diagonal black line represents perfect forecasts.}
\label{child_cal}
\end{figure}

Figure \ref{child_cal} shows a calibration plot for both procedures at sample sizes 100, 500, and 1000. As with the Hepar2 simulation, overestimating actual performance is more common than understimating performance.

\section{Discussion and conclusion}
\label{discussion}
\subsection{Implications}

In carrying out these simulations, we hoped to begin answering some important questions about resampling in the context of causal discovery. For example, when it comes to estimating the causal graphical model of an unknown causal data generating process, should we prefer off-the-shelf causal discovery methods, or a bootstrap ensemble, or a jackknife ensemble? These simulations suggest that bootstrapping might be preferred, and that the difference between all three options may not be large, at least for the GES algorithm. It also appears to be the case that this question does not have a simple answer: the preferred method might depend on sample size and the difficulty of the task.

Another question is: Should we prefer bootstrapping or jackknifing as indicators of confidence in discovered graph features? As with accuracy, it again appears that the preferred method might depend on sample size and the difficulty of the task. Overall, though, in these simulation bootstrapping had better calibration than jackknifing in most cases.

Finally, can resampling be considered a reliable way to evaluate confidence in discovered graph features? These simulations indicate that in some circumstances it can be. Bootstrapping had very good reliability for both the linear Gaussian and Hepar2 simulations, which is impressive given that the data generating models used in these two simulations are quite different. While its performance on the Child simulation was far from what one would want, the plot suggests that it may improve at higher sample sizes.

\subsection{Limitations and future directions}

Our findings have a number of limitations. First, they may not generalize to other search algorithms, such as PC or FCI \citep{spirtes2000causation}, as we only tested the GES algorithm. In our limited experience PC performs substantially better when bootstrapped, but we have not carried out a complete simulation study to confirm this. It is also unclear how well resampling on causal discovery algorithms that consider latent variables performs, although some results on that problem has been produced by \citep{fattaneh}. These areas provide a natural extension to the work presented in this paper, and would be useful for some of our current work in medical science domains \citep{anker2019causal}

Our findings also may not apply to real world problems that are very different from the simulated models and expert models evaluated here, or to data sets with sample sizes outside the range of those we tested. We only looked at sample sizes in the range of 100 to 1000. Sample sizes below 100 should probably be analyzed with caution under any circumstances, but there are many real world data sets with sample sizes much larger than 1000.

We also only considered a small subset of possible kinds of graphical structures. The linear Gaussian simulations were fairly sparse, and we only considered two expert models, which have invariant causal structure. Our findings may not apply to data generating models with graphs that are very dense or very sparse, with small-world graphs, or with graphs structurally different than those we used in our simulations in other ways.

None of our simulations included models with more than 100 variables. As such, our results may not apply to data with thousands of variables or more. We also did not consider continuous variable models with non-Gaussian noise or nonlinear relationships, so our results may not apply to data-generating models that have non-Gaussian noise or nonlinear relationships.

Overall, it is still unclear at this time why calibration may have good performance in some circumstances and bad performance in other circumstances. If resampling is not always well calibrated, we should identify features, ideally identifiable from the data themselves or from easily accessible meta-data, that determine the calibration of these resampling methods. Extending these simulations to other algorithms and other data-generating models will help us to understand what those features might be.

\section*{Acknowledgment}

The authors would like to thank the Center for Causal Discovery for developing and supporting the open source Tetrad software package.

\bibliographystyle{acm}
\bibliography{bibliography}

\end{document}